# Explaining Classifications to Non-Experts: An XAI User Study of Post-Hoc Explanations for a Classifier When People Lack Expertise

Courtney Ford[1,3] and Mark T. Keane[1,2]

[1]    School of Computer Science, University College Dublin, Dublin, Ireland
[2]    Insight Centre for Data Analytics, UCD, Dublin, Ireland
[3]    ML Labs, SFI CRT, Ireland
                    courtney.ford@ucdconnect.ie, mark.keane@ucd.ie

**Abstract.** Very few eXplainable AI (XAI) studies consider how users' understanding of explanations might change depending on whether they know more/less about the to-be-explained domain (i.e., whether they differ in their expertise). Yet, expertise is a critical facet of most high-stakes, human decision-making (e.g., understanding how a trainee doctor differs from an experienced consultant). Accordingly, this paper reports a novel, user study (N=96) on how people's expertise in a domain affects their understanding of post-hoc explanations-by-example for a deep-learning, black-box classifier. The results show that people's understanding of explanations for correct and incorrect classifications changes dramatically, on several dimensions (e.g., response times, perceptions of correctness and helpfulness), when the image-based domain considered is familiar (i.e., MNIST) as opposed to unfamiliar (i.e., Kannada-MNIST). The wider implications of these new findings for XAI strategies are discussed.

## 1    Introduction

As Artificial Intelligence (AI) extends its automated decision-making capabilities to tasks that impact our everyday lives, there is an urgency to solve the problem of explaining the operation of these systems to human end-users using Explainable AI (XAI). Furthermore, opaque deep learning models are increasingly used for high-stakes decision-making tasks (e.g., cancer detection [9] and crime occurrence [15]); decisions that need to be explainable to audiences with varying levels of expertise. However, very little hard evidence exists on whether people with different levels of domain knowledge understand automated explanations differently. In this paper, we consider how *domain expertise* changes people's understanding of automated, post-hoc example-based explanations of a deep learner's classifications of images of written numbers with which users have differential familiarity (i.e., MNIST and Kannada-MNIST).

## 1.1    Post-Hoc Explanation of Unfamiliar Images

Consider a medical scenario where a black-box classifier analyzes MRI images for cancer detection. The system might explain a detection of a cancerous mass by showing end-users similar historical images with the same diagnosis (i.e., socalled post-hoc, example-based explanations [20]). Alternatively, it might highlight a critical region in the MRI to show the key features leading to the diagnosis (i.e., feature-importance explanations [26,23]. Although many XAI methods generate such explanations, there is a dearth of user studies on how people understand these explanations and whether they really work as intended (e.g., see [16]). Even fewer user studies have considered whether different users who have different expertise in a domain, understand automated explanations differently.

However, intuitively, expertise is likely to be important to people's understanding of explanations. For instance, a medical consultant with many years' experience will engage very differently with an MRI image from a student doctor. The consultant may "read" the image quicker, find critical features more readily, and reach different diagnostic conclusions based on their deeper expertise. So, automated explanations may need to consider expertise differences in what the model presents to users to support decision making. Unfortunately, most XAI methods assume a one-size-fits-all approach to explanation, that all users should receive the same explanation irrespective of their background. In this paper, we present a study that reveals wholly-new evidence on how users' perceptions of explanations can change based on their expertise with an image-based domain analyzed by an AI classifier.

The present study examines post-hoc, example-based explanations for classifications made by a convolutional neural network (CNN) involving images of written numbers (from the MNIST and Kannada-MNIST datasets). Kenny et al. [19,20] have previously shown that correct and incorrect MNIST classifications by this CNN can be explained by finding nearest-neighboring examples to a query-image from the training data, using their twin-systems XAI method (see Figure 1). These example-based explanations give users insight into why the model classified an image in a certain way (see Figure 2). In a series of user studies, Kenny et al. [18] showed that people who were given these explanations judged misclassifications differently to those who did not receive explanations. Specifically, they found that explanations led people to judge misclassifications as being more correct, even though they still recognized them as incorrect (n.b., similar effects did not occur in correct classifications). They also showed that people experienced algorithmic aversion; users trust in the model decreased when they were shown more and more misclassifications. However, these results involved written Arabic numerals commmonly used by English-speaking participants (i.e., the MNIST dataset), a well-learned domain from an early age, regularly practiced when reading handwritten information. So, everyone was an deep-expert with these written-numbers image-data. This paper examines how a Western, English-speaking population responds to example-based explanations from an unfamiliar, Indian number system (i.e., the Kannada-MNIST dataset). By performing matched tests with explanations of the CNN's classifications for these two different written-numeral systems, we aim to assess cognitive differences that arise from differences in expertise.

We report a large user study examining the effect of domain expertise on example-based explanations for classifications made by a CNN. The key manipulation used to test for expertise effects involved presenting participants with familiar (i.e., Arabic numerals from the MNIST dataset) or unfamiliar (i.e., Indian numerals from the Kannada-MNIST dataset) written numbers. In the following sub-section, we quickly describe the main differences between these two image datasets (see 1.2), before outlining the structure of the paper (see 1.3).

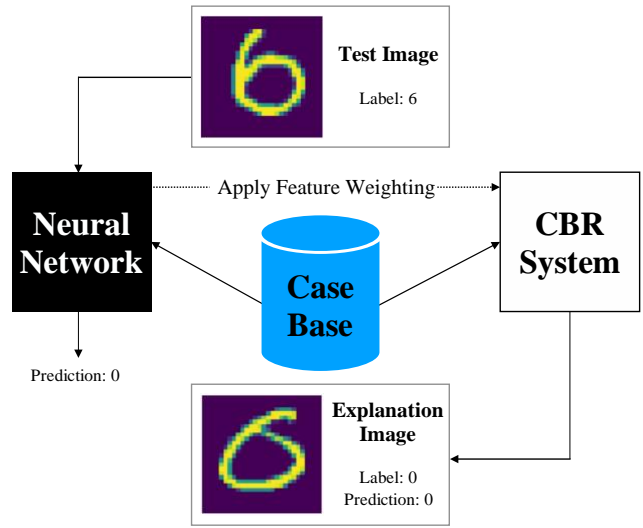

**Fig.1.** Twin-Systems XAI (Adapted from [19,20]): Feature weights from a CNN classifier are mapped to a twinned Case Based Reasoner (i.e., a *k*-NN), so the classification of the test image is explained post-hoc by close examples from the dataset/case-base.

## 1.2   A Tale of Two Written-Number Systems

The MNIST dataset consists of 70,000 images of handwritten, single-digit Arabic numerals. MNIST is a ubiquitous machine learning dataset; its simplicity and "too perfect" data features (i.e., the images uniformly centered with the digits of equal sizes) routinely results in accuracies >99% for many models [1]. MNIST has also been heavily explored in the XAI literature, often user-tested in a "classifier debugging" task of the sort used here (e.g., see [18,10,28]).

Several newer datasets address MNIST's limitations. One such dataset, the Kannada-MNIST [25], is comprised of ten classes of data (representing Arabic numerals 0-9) in the Kannada script. The Kannada language is spoken in Karnataka, India, and its single-digit numeral system is comparable to MNIST in its relative simplicity (i.e., most characters are single-stroke). Over 200 published AI papers have used the Kannada-MNIST dataset since 2019; however, to our knowledge, no XAI studies using the dataset have been completed.

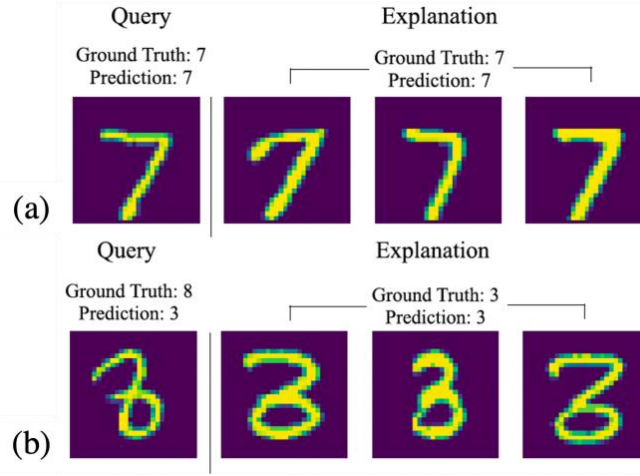

**Fig.2.** Examples of post-hoc, example-based explanations for (a) a correct and (b) incorrect classification. The query image gives the CNN's prediction, along with three nearest neighbours in the dataset that influenced the classification made [19,20].

Figure 3 shows samples of the MNIST and Kannada-MNIST datasets. These different written-numeral datasets give us an intuitive sense of how expertise might impact people's understanding of them. Readers familiar with the symbol "two" in Figure 3a understand that the first two symbols, beside the printed number (i.e., in the blue box), can be a "two" even though the second contains a loop feature absent in the others. We also know that even though an "eight" has a similar loop, it is a different number-class to "two", because it has a closed upper-loop that the "two" lacks. So, expertise with these written numbers allows us to judge "acceptable variations" in the class instances and judge the class boundaries between classes when progressive changes make it a different number.

Now, consider the unfamiliar Kannada dataset, assuming you are an English-speaking Westerner with no knowledge of Indian languages. Figure 3b shows the Kannada symbol "mu͞ru" (the symbol for the number 3). Without domain expertise, it is hard to judge whether the three symbols to the right of the printed number (i.e., in the blue box) are valid variants of "mu͞ru" or whether they are from different number-classes. In fact, the third symbol (in the orange box) is from a different number-class, though it is very similar. As in MRI interpretation by doctors, non-expert users may not easily determine which classifications are correct or incorrect. Non-experts may not notice subtle differences between images and, indeed, may find explanations harder to understand. It is these aspects of the user experience that we explore in the present study.

### (a) MNIST

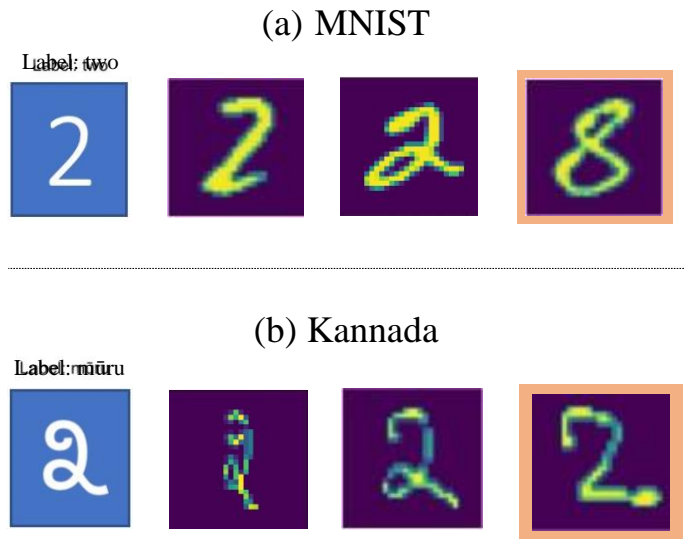

### (b) Kannada

**Fig.3.** Examples of variation within a class for the (a) MNIST and (b) Kannada datasets. Images outlined in orange are numbers from different, but similar, classes.

## 1.3    Outline of Paper

The following section outlines related work on expertise and how it might impact AI systems (see section 2). In the remainder of the paper, we report a user study that examines the effect of domain expertise and example-based explanations on people's judgments of a CNN's correct and incorrect classifications using the MNIST and Kannada datasets (see sections 3, 4, and 5). The study tests the local effects of domain expertise using a comparison of the MNIST versus Kannada-MNIST (henceforth, Kannada) dataset. This research is wholly novel in the current XAI literature (see section 2) and reveals how expertise impacts several different psychological measures. In section 6, we consider the broader implications of these findings for the interpretability of black-box models.

## 2    Related Work

Experts are individuals who have superior knowledge of a domain, and their expertise is generally assumed to be acquired through learned experience [36]. Evidence from cognitive science suggests that experts profit from perceptual processes that aid in the storage, retrieval, and analysis of domain-familiar stimuli. One such process called "chunking" enables experts to store units of information in a single, compact group rather than as separate individual entities [6,5,12]. Though experts can exploit chunking for goal-oriented tasks (e.g., expert chess players use goal-oriented chunking to recall and anticipate moves more clearly and quickly than less experienced players [7]), chunking is primarily a subconscious, perceptual process learned from experience [11]. For example, readers and writers of languages with complex character systems (e.g., Chinese) perceptually chunk "semantic-phonetic" compound characters as a cohesive unit rather than separate parts; a skill that develops gradually alongside improving language expertise in vocabulary and grammar [2]. Other aspects of perceptual expertise, such as visual search, also appear to develop over time. Krupinski et al. [22], in a longitudinal study, found that resident pathologists became increasingly more efficient at viewing domain-familiar images and focusing on key image details over the term of their residencies.

There is a considerable literature on the impact of domain expertise on people's perceptual processing of images. Consistent findings suggest that experts outperform non-experts in reaction time [34], visual span [34,4], and change detection [35]. In a series of user studies, Thompson and Tangen [30] manipulated the amount of perceptual information (i.e., image quality and time) available to fingerprint experts and novices. The experts outperformed novices in measures of accuracy and time despite image noise and temporal spacing; their results suggested that experts benefit from "non-analytic" processing of domain-familiar data that supersedes information quality. However, this advantage disappears for non-domain-related images [17]. Searston and Tangen [29] found that domain experts have an accuracy and speed advantage over novices when identifying matches and outliers in familiar stimuli (i.e., fingerprint images), but show the same behaviour as novices when identifying non-domain images (i.e., inverted faces). So, the improved perceptual abilities and task performance of experts only holds when the domain is familiar to them; experts and non-experts perform the same when the domain is unfamiliar to both groups. To that end, it is possible that non-experts could benefit from feature signaling (e.g., highlighted data features [21]), or by presenting tasks as similarity judgments [27].

The wealth of cognitive research on expertise suggests that it influences analytic and decision-making tasks. But, what if analyses and decisions are made by an AI system? Recently, a small number of papers have directly investigated the role of domain expertise on trust in an AI system. Nourani, King, and Ragan [24] report a user study on the role of expertise and AI-system first-impressions on users' trust. They found that domain experts are sensitive to system error based on their first impressions of an AI system; experts who saw correct classifications at the start of the system interaction reported higher trust in the overall system than experts who first viewed system errors. Novices, on the other hand, were not sensitive to material-type order, but over-estimated the system's accuracy despite viewing multiple misclassifications. Bayer, Gimpel, and Markgraf [3] found that explanations of a decision-support system contributed to higher trust in the overall system for domain experts but failed to improve trust in non-experts. It appears that experts augment the system's explanation with their domain knowledge, whereas the same explanation has little significance to someone with no or limited domain understanding. Dikman and Burns [8] reported a user study in which participants (all of whom were non-experts) were provided with an occasionally faulty AI Assistant for an investing task; the experimental condition additionally received domain information that stated why a given factor influences the decision. They found that participants made fewer risky investments when provided with domain information, sometimes in opposition to the AI assistant. In fact, these participants reported less trust in the AI system [8]. Therefore, it appears that augmenting expertise throughout a user-system interaction improves user performance at the expense of trust if the system is repeatedly inaccurate or provides faulty recommendations.

The literature described above indicates fundamental perceptual differences between experts and non-experts that impact the effectiveness of XAI explanation strategies. Experts process familiar information faster and more accurately than novices, though this advantage disappears in unfamiliar domains. Therefore, some explanation strategies may slow experts down (i.e., if the data is familiar), while others assume a higher level of knowledge or experience than the user possesses. Experts also seem to be sensitive to domain-familiar system errors, an issue possibly resolved by an alternate, error-specific explanation strategy. Non-experts, on the other hand, may require an alternative explanation strategy to experts. Though novices generally perform worse than experts in the experts' domain, consistent findings show that novices are just as accurate as experts in unfamiliar domains. From an XAI perspective, these findings suggest that explanations for novices could close the accuracy gap through supplemental information or guided assistance. Recently, Herchenbach et al. [13] have argued that the explanation strategy for experts using these domains should use classspecific prototypes, near misses (i.e., counterfactuals) and near hits (i.e, semifactuals) and show how these can be computed. However, their proposals are not grounded in any user analysis and some caution is warrented in juggling so many different explanation options [33,32]. The current study explores the impact of domain expertise on the efficacy of post-hoc, example-based explanations by varying the familiarity of the domains presented to a single group.

## 3     User Study: Expertise in Image Classification

Kenny et al. [18] user tested post-hoc example-based explanations of a CNN's classifications of image data (i.e., MNIST data). They extracted correct and incorrect classifications from a CNN, explaining them with nearest neighbours using their twin-systems method. In a "classifier debugging" task, participants judged the correctness of classifications with and without explanations and it was shown that example-based explanations changed people's perception of misclassifications (but not correct classifications); people given explanations viewed these errors as "less incorrect" than those not getting explanations. So, explanations partly mitigated people's perception of errors made by the AI system.

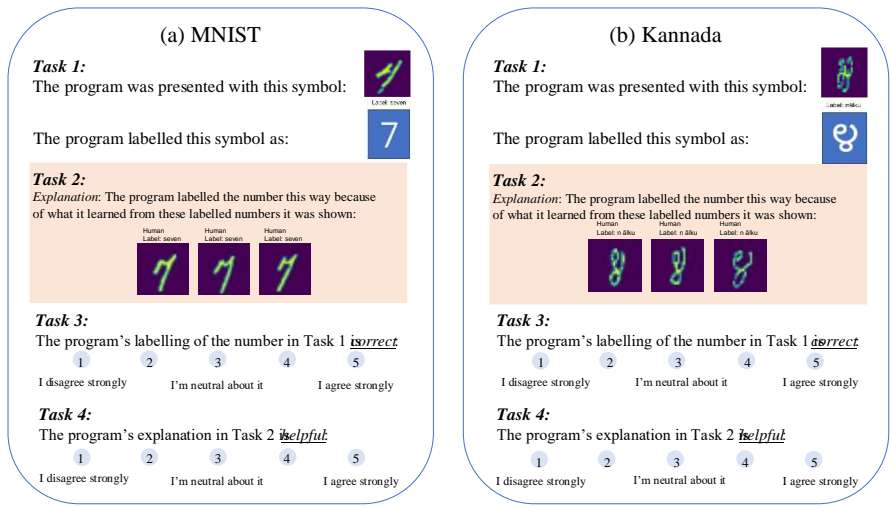

**Fig. 4.** Sample materials used in the study showing (a) MNIST and (b) KannadaMNIST misclassifications along with explanations showing three nearest neighbours.

However, these tests were performed on a familiar dataset (the Arabic numerals of MNIST). At present, no evidence in the literature exists on how people respond to an unfamiliar, image dataset (such as the Kannada-MNIST one).

### 3.1     Method

The present study compares people's performance on the MNIST dataset to a matched set of images from the Kannada dataset, using the measures used before (Correctness, Trust, Satisfaction) and several new measures (Helpfulness and Response Time). In the study, people were presented with matched number-images from the MNIST and Kannada datasets and post-hoc, example-based explanations. Participants then judged the correctness of the model's classifications and provided other subjective judgments (e.g., helpfulness of explanations and trust and satisfaction in the overall system after each dataset). Figure 4 shows sample materials as they were presented to participants in the study.

**Participants** (N=96) were recruited on the Prolific crowdsourcing site. Eligibility criteria were that the participants had to be over 18 years old, native English speakers, and residents in the USA, UK, or Ireland. Participants were randomly assigned to the conditions of the study. These chosen N was based on a power analysis for a moderate effect size. Ethical clearance was provided by the University's Ethics Panel (LS-E-21-214-Ford-Keane).

**Materials** used were the actual classifications (correct and incorrect) produced by a CNN trained separately on MNIST and Kannada-MNIST datasets. The model has a high accuracy for both domains (~96%); therefore, multiple runs were used to produce materials. The misclassifications were system mislabels to a class different from the ground-truth label. The image-instances used in the experiment for the classifications consisted of six number classes (1, 3, 4, 5, 6 in Arabic numerals; 2, 4, 5, 7, 8, 9 in Kannada). These selected classes were chosen on the need to match material sets between the domains, a procedure that required time and effort. Materials across and within the datasets were matched using the Structural Similarity Index Metric[1] [31]; t-tests revealed no significant difference between the MNIST and Kannada materials chosen. Explanatory nearest-neighboring examples (3 were always provided) were found using the twin-systems method, which is an accurate feature-weighting method for finding example-based explanations for CNN classifiers (see [19,20]. The study used 42 distinct materials with equal numbers from each dataset and class (21 each; 12 were correct and 9 were misclassifications). So, the error-level for the study was high at 43%.

**Procedure, Measures & Analyses** After being told the system was a program that "learned" to classify written symbols, participants were told that they would be shown several examples of its classifications. Instructions informed participants that they would see two different groups of symbols (the MNIST and Kannada symbols) in two different sections of the study (i.e., two distinct blocks of items counterbalanced across groups for order). Before beginning each section, participants were provided with the centroid images and written labels of each number-class of the dataset being tested. They were also told that these symbols could be written in different ways. For each presented item, they were told that their task was to rate the correctness of the presented classification on a 5-point

---

[1] SSIM distances were calculated from each test instance to its centroid, and its three explanatory cases to their respective centroid.

Likert-scale from "I disagree strongly" (1) to "I agree strongly" (5), as well as the helpfulness of the system's explanation on another 5-point Likert scale (see Figure 4 for sample materials). After rating all of the presented classifications, participants filled out Hoffman et al.'s [14] DARPA trust (8 questions) and satisfaction (8 questions) surveys. The 4 measures analyzed were:

- *Correctness*. Mean 5-point Likert-scale ratings of perceived correctness of correct system classifications and misclassifications.
- *Helpfulness*. Mean 5-point Likert-scale ratings of perceived explanation helpfulness of correct system classifications and misclassifications.
- *Response Time*. Time (in seconds) spent on each page of the experiment (i.e., each material item), including viewing the presented image and explanatory images, and performing Correctness and Helpfulness judgements.
- *Trust and Satisfaction*. Ratings from the DARPA Trust and Satisfaction surveys analyzed question-by-question. This variable was repeated after each dataset (MNIST and Kannada datasets).

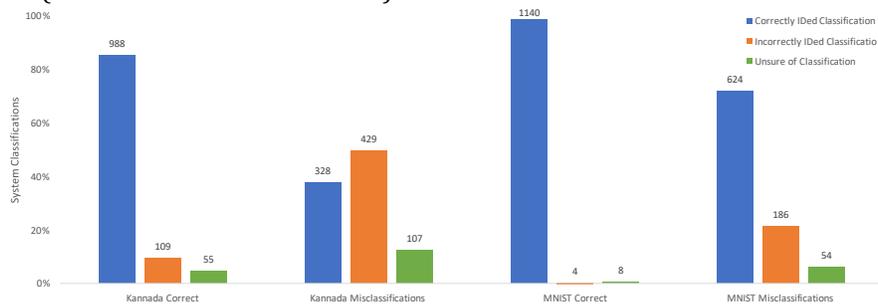

**Fig. 5.** Percent of user identifications by type-of-classification. Scores from the Correctness ratings were re-coded categorically as "identified-as-correct" (if rated ≥ 4 on the correctness scale) or "identified-as-error" (if rated ≤ 2 on the correctness scale).

**Design** was a 2 (Order: MNIST-first, or Kannada-first, a between-subjects variable) x 2 (Dataset: MNIST or Kannada, a within-subjects variable) x 2 (Classification Type: correct/incorrect classification, a within-subjects variable).

### 3.2    Results & Discussion

The results of this experiment show that people are profoundly affected by their lack of expertise with an unfamiliar domain relative to a familiar one. When interacting with an unfamiliar domain (the Kannada dataset), participants experienced greater difficulty accurately identifying correct and incorrect classifications. Their judgments of system correctness and explanation helpfulness also changed, and response times were longer. Non-experts can perceptually compare a written-number image to its ideal-printed image of the number class, irrespective of the domain. However, they have limited background knowledge about how these written numbers can vary within the labeled number-class before transitioning to another number-class (hence, their judgmental uncertainty)[2].

**Correctness** The three-way mixed measures ANOVA revealed a significant two-way interaction between Dataset and Classification Type, $F(1,94) = 310.58$, $p< 0.001$, partial $\eta^2 = 0.77$. Simple main effects of Dataset revealed that participants judged correctly-predicted MNIST predictions as significantly more correct ($M=4.8$, $SD=0.3$) than the correctly-predicted Kannada items ($M=4.26$, $SD=0.6$, $p < 0.001$). Conversely, participants rated Kannada misclassifications as significantly more correct ($M = 3.15$, $SD=0.82$) than MNIST misclassifications ($M = 2.15$, $SD=0.68$, $p< 0.001$). So, when people are less familiar with a domain, they are less sure about judging the correctness of items; for the Kannada data, judgments of correct and incorrect classifications move towards the center of the 5-point scale reflecting this uncertainty.

Participants were also less sure about correct and incorrect items in the unfamiliar domain relative to the familiar one (see Figure 5). This effect is revealed when we re-code their correctness ratings categorically as "identified-as-correct" (if rated ≥ 4 on the correctness scale) or "identified-as-error" (if rated ≤ 2 on the correctness scale). Then, using this re-coding we can show how many item-classifications in each dataset were correctly or incorrectly identified by participants. Figure 5a graphs this data and shows how people's performance on the two datasets differs. For the MNIST dataset items that were correctly classified by the model are correctly identified (∼100% of the time, 1140 out of 1144), but for the Kannada dataset they are only identified (90% of the time, 988 out of 1097). This effect is even more marked for misclassifications. For the MNIST dataset, items that were incorrectly classified by the model (i.e., misclassifications) are correctly identified (77% of the time), but for the Kannada dataset they are only correctly identified (43% of the time). So, people encounter some difficulty spotting correctly-classified items and have major difficulties identifying incorrectly-classified items in the unfamiliar Kannada images. An *a posteriori* chi-square test of this data found a significant association between Dataset and participants' correct identification of the CNN's classifications as either correct or incorrect, $\chi^2(1) = 37.71$, $p< 0.0001$; an association of $\phi = 0.11$ ($p< 0.0001$).

**Helpfulness** People's rating of how helpful they found the system explanations also varied for the familiar versus unfamiliar datasets. A three-way mixed measures ANOVA analysis of Helpfulness revealed a significant two-way interaction of Dataset with Classification Type, $F(1, 94) = 51.9$, $p < 0.01$, partial $\eta^2$ = 0.36. Simple main effects revealed a counter-intuitive result: that nearestneighbor example explanations were perceived as more helpful when judging correct classifications for MNIST items ($M = 4.77$, $SD = 0.39$), compared to correct classifications for Kannada items ($M = 4.37$, $SD = 0.63$, $p < 0.001$). Furthermore, for the MNIST items alone, these explanations were perceived as more helpful when provided for correct classifications ($M = 4.77$, $SD = 0.39$) compared to misclassifications ($M = 3.49$, $SD = 0.87$, $p < .001$). The same pattern is found within Kannada items with simple main effects showing that explanations were perceived as more helpful when provided for correct classifications ($M = 4.37$, $SD = 0.63$) than for misclassifications ($M = 3.71$, $SD = 0.77$). These results suggest that examples might not be appropriate for explaining classifications in unfamiliar domains, as users' perceptions of helpfulness seem to follow ease-with-the-task. People may simply think explanations are more helpful just because they can readily identify them as correct.

---

[2] Note, the Order variable is not reported as it had no effect on the analyses reported; that is, the presentation of one dataset before the other did not affect performance.

**Response Time** The differences caused by expertise were also seen in the mean response times to items in the different conditions[3]. A three-way mixed measures ANOVA revealed a significant two-way interaction of Dataset with Classification Type, $F(1, 94) = 10.1$, $p < 0.01$, partial $\eta^2 = 0.097$. Simple main effects revealed significantly longer response times for correct Kannada materials ($M = 13.26$, $SD = 7.79$) than for correct MNIST materials ($M = 11.54$, $SD = 4.84$, $p < .01$; see Figure 6). For the MNIST dataset, response times were significantly shorter for correct classifications ($M = 11.54$, $SD = 4.82$) than misclassifications ($M = 15.38$, $SD = 6.19$, $p < 0.001$). Similarly, for Kannada dataset, response times were significantly shorter for correct classifications ($M = 13.26$, $SD = 5.79$) than misclassifications ($M = 15.43$, $SD = 6.94$, $p < 0.001$). When these results are combined with the accuracy differences in users identification of correct and incorrect classifications, it appears that longer response times improve users' accuracy in "identifying-as-correct" correct classifications of unfamiliar data. Conversely, users spent approximately the same time analyzing MNIST and Kannada misclassifications, despite struggling to "identify-as-error" the system misclassifications of unfamiliar data. These results suggest that there may be a speed versus accuracy trade-off occurring for misclassifications.

**Trust & Satisfaction** Repeated-measures ANOVA analyses of Trust and Satisfaction (respectively) provided no significant effects across the conditions, but the questions differed significantly within each questionnaire. There is insufficient space to discuss the significance of these measures here.

---

[3] Outlier responses (3 *SD*s from the mean for a given item) were removed before mean times were analyzed.

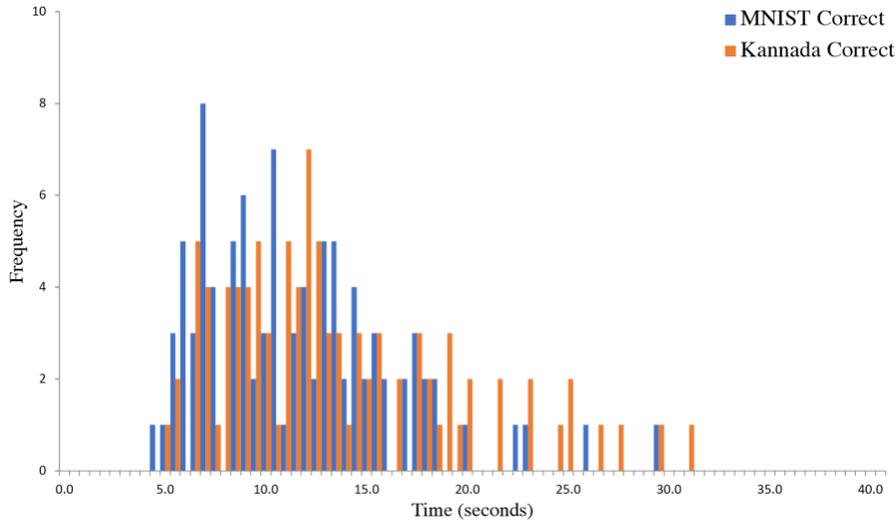

**Fig.6.** Distribution of response time (in seconds) for correctly classified MNIST and Kannada materials.

## 4   General Discussion

The present study reveal several novel findings on the impact of domain expertise on people's perception of a classifier's decisions when post-hoc, example-based explanations are provided. We found that (i) domain familiarity significantly impacts users judgments of a classifier, (ii) example-based explanations assist users with judging correct system classifications of an unfamiliar domain, but (iii) people have significant difficulties dealing with misclassifications in an unfamiliar domain (uneased by the provision of explanations).

This study determines the impact of domain familiarity on user judgments of an AI system's classifications. As expected, users displayed an accuracy and speed advantage in their judgments of correct classifications for the familiar domain over the unfamiliar one. Users were slower (albeit still accurate) at positively identifying correct, unfamiliar classifications but notably struggled with judging and identifying unfamiliar misclassifications. In fact, they tend to perceive Kannada misclassifications as being correct (and making this decision quickly). These results echo those of Searson and Tangen [29], who found that experts are adept at identifying matches and outliers in domain-familiar data and consistently have an accuracy and speed advantage over non-experts. This ability may stem from chunking variations for a familiar class; in other words, experts may be perceptually aware of the bounds of a familiar class. This suggests that modifications to an explanation strategy should address the perceptual limitations of non-experts (e.g., by providing information about class variation).

The study also shows how post-hoc, example-based explanations impacted non-experts' performance in judging a black-box classifier. Manifestly, we found that explanations did not improve user accuracy in identifying misclassifications in the unfamiliar domain. It may be that example-based explanations improve non-experts' accuracy for within-class, but not between-class, variation. Despite this limitation, users who received an explanation reported it as being helpful (suggesting that in this context the explanation can be misleading). We also explored the effect of domain expertise on user trust in a system and investigated whether post-hoc, example-based explanations mitigated or improved this effect. Surprisingly, expertise did not change users' trust in the system.

With respect to future research the findings of the current study highlight an avenue for a modified explanation strategy to address the unique challenges of unfamiliar domains. We have seen that users struggle, in particular, with correctly identifying and judging system misclassifications with unfamiliar data. Future research focusing on improving people's performance, through an novel explanation strategy catering for domain expertise may improve people's perceptions of the system as a whole. Such a solution should increase the interpretability of deep learning systems. In conclusion, the present studies present a rich set of findings for a wider understanding of the dynamics of user interactions between explanation-strategies and domain expertise in XAI contexts.

**Acknowledgements.** This paper emanated from research funded by Science Foundation Ireland to the Insight Centre for Data Analytics (12/RC/2289P2) and SFI Centre for Research Training in Machine Learning (18/CRT/6183).